\documentclass[letterpaper, 10 pt, conference]{ieeeconf}  

\IEEEoverridecommandlockouts                              

\usepackage{amsmath} 
\usepackage{amssymb}  
\usepackage{graphics} 
\usepackage{graphicx}
\usepackage{caption}
\usepackage{subcaption}
\usepackage{amsmath} 
\usepackage{amssymb}  
\usepackage[table]{xcolor}
\usepackage[font=small,belowskip=0pt, aboveskip=6pt]{caption}
\usepackage{multirow}
\usepackage{url}
\usepackage{booktabs,amsfonts,dcolumn}
\usepackage{tabu}
\usepackage{gensymb}
\usepackage{makecell}
\usepackage{hyperref}
\usepackage{comment}
\usepackage{xspace}
\usepackage{tabularx}
\usepackage{import}
\let\labelindent\relax
\usepackage{enumitem}

\usepackage{hyphenat}
\usepackage[group-separator={,}]{siunitx}
\usepackage[style=ieee,mincitenames=1,maxcitenames=2,maxbibnames=10,doi=false,isbn=false,url=false,eprint=false]{biblatex}
\DeclareSourcemap{
  \maps[datatype=bibtex, overwrite]{
    \map{
      \step[fieldset=address, null]
      \step[fieldset=location, null]
    }
  }
}
\include{macros}

\urldef{\mailsb}\path| yiannisr@cse.sc.edu|

\long\def\invis#1{}

\newcommand\eq[1]{Eq.~\eqref{#1}}
\newcommand\fig[1]{Fig.~\ref{#1}}

\newcommand\tab[1]{Table~\ref{#1}}

\makeatletter
\DeclareRobustCommand\onedot{\futurelet\@let@token\@onedot}
\def\@onedot{\ifx\@let@token.\else.\null\fi\xspace}

\def\etal{\emph{et al}\onedot}

\makeatother

\title{\LARGE \bf High Definition, Inexpensive, Underwater Mapping}
\author{ Bharat Joshi, Marios Xanthidis, Sharmin Rahman, Ioannis Rekleitis%
\thanks{The authors are with the Computer Science and Engineering Department, University of South Carolina, Columbia, SC, USA, 29208, {\tt\small \{bjoshi,mariosx,srahman\}@email.sc.edu,  yiannisr@cse.sc.edu.}}%
\thanks{
This research has been supported in part by the National Science Foundation under grants 1943205 and 2024741. The authors would also like to acknowledge the help of the Woodville Karst Plain Project (WKPP) and El Centro Investigador del Sistema Acuífero de Quintana Roo A.C. (CINDAQ) in collecting data, providing access to challenging underwater caves, and mentoring us in underwater cave exploration. Last but not least, we would like to thank Halcyon Dive Systems for their support with equipment.}%
}

\begin{document}

\begin{minipage}{0.95\textwidth}\ \\[12pt]  
\begin{center}
     This paper has been accepted for publication in \textit{IEEE Conference on Robotics and Automation 2022}.  
\end{center}
  \vspace{1in}
  ©2022 IEEE. Personal use of this material is permitted. Permission from IEEE must be obtained for all other uses, in any current or future media, including reprinting/republishing this material for advertising or promotional purposes, creating new collective works, for resale or redistribution to servers or lists, or reuse of any copyrighted component of this work in other works.
\end{minipage}

\newpage

\maketitle              

\thispagestyle{empty}
\pagestyle{empty}

\begin{abstract}
In this paper we present a complete framework for Underwater SLAM utilizing a single inexpensive sensor. Over the recent years, imaging technology of action cameras is producing stunning results even under the challenging conditions of the underwater domain. The GoPro 9 camera provides high definition video in synchronization with an Inertial Measurement Unit (IMU) data stream encoded in a single mp4 file. \invis{Quite often, action cameras are attached on autonomous or remotely operated underwater vehicles (AUV/ROV), or sensors suites providing visual testimony of the experiment. With the proposed framework, the visual/inertial data from the action camera can be integrated with the data from the AUV/ROV in post processing to enhance the accuracy of the trajectory and the visual resolution of the map. Two different synchronization approaches between the stand\hyp alone GoPro and the other sensors are examined. Furthermore, the Visual/Inertial SLAM framework is augmented with global orientation data from an external magnetometer.} The visual inertial SLAM framework is augmented to adjust the map after each loop closure. Data collected at an artificial wreck of the coast of South Carolina and in caverns and caves in Florida demonstrate the robustness of the proposed approach in a variety of conditions. 
\end{abstract}

\section{Introduction}
The underwater domain presents a special allure since the early days of exploration~\cite{cousteau1953silent}; coral reefs, shipwrecks, and underwater caves all present unique views like nothing most people see above water. Underwater exploration using acoustic sensors is well studied, however, the resulting representations convey only limited information; in contrast vision based mapping presents the most familiar representations~ \cite{eustice2006visually,demesticha20144th,gonzalez2014catlin,ModasshirICRA2020}. Underwater is a very challenging environment for cameras.  The visibility is limited, sometimes objects after a few meters disappear; color attenuation, colors disappear with depth starting with red~\cite{SkaffBMVC2008,roznere2019real}; floating particulates generate blurriness; there is varying illumination resulting from caustic patterns due to waves up to complete lack of ambient light inside caves; and the reduced number of features  makes localization challenging. 

Autonomous Underwater Vehicles (AUVs) and Remotely Operated Vehicles (ROVs) range in cost from a few thousand to hundreds of thousand of dollars. Furthermore, camera technologies for these vehicles, unless at the higher end of the spectrum provide images not of the highest quality; one major challenge is the light has to pass from water to the AUVs window, through air, then the lens of the camera generating additional distortions. In the recent years so\hyp called action cameras and in particular the GoPro cameras have produced exceptional imagery for a fraction of the cost. The improvements in image quality though, were limited by the single camera view which made estimation of scale near impossible. As shown in Joshi \etal~\cite{JoshiIROS2019} and Quattrini Li \etal~\cite{QuattriniLiISERVO2016} monocular vision without inertial data has very low accuracy. From the GoPro 5 black, the video contains embedded inertial data at a rate of 200 Hz, without synchronization information. Starting from GoPro 8, the video contains inertial information along with necessary timing information for camera IMU synchronization thus making underwater state estimation feasible. In this paper we tested some of the most promising open\hyp source Visual Inertial SLAM packages~\cite{RahmanIROS2019a,vinsmono,orbslam3,openvins,colmap_sfm}, in a variety of environments with very accurate results. Furthermore, the SVIn2 framework~\cite{RahmanIROS2019a} is augmented with updating the 3D pose of the detected visual features after loop closure producing a consistent global map. The code is publicly available\footnote{\url{https://github.com/AutonomousFieldRoboticsLab/gopro_ros}}.  

\begin{figure}[t]
\centering
\includegraphics[height=0.17\textheight]{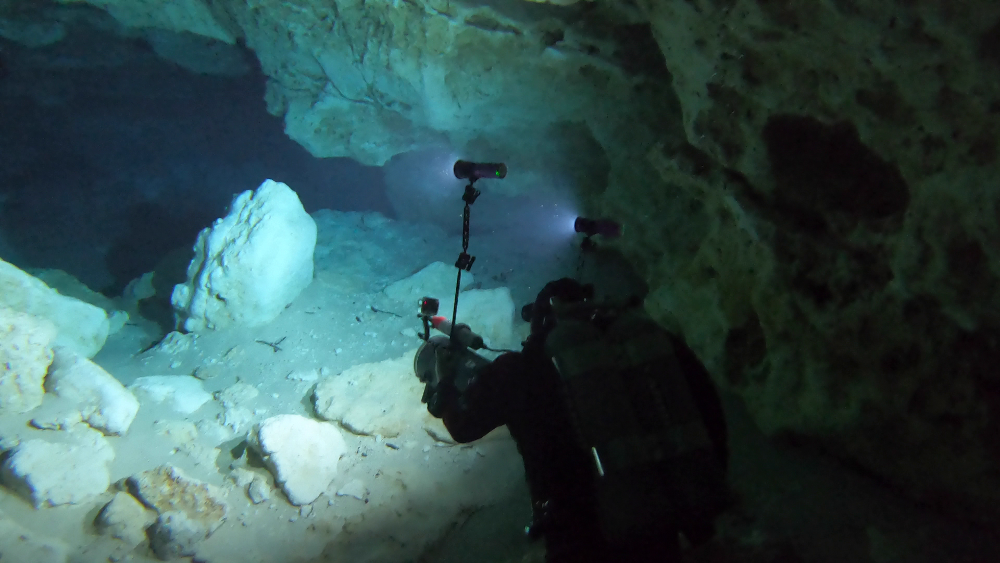}
\caption{Collecting data in an underwater cavern, Ginnie Spring, FL, USA. GoPro 9 camera is attached to the stereo\hyp rig~\cite{RahmanOceans2018}, lighting from two Keldan lights~\cite{Keldan}.}
\label{fig:SpeedoStav}
\vspace{-0.1in}
\end{figure}
\invis{
The GoPro cameras are self contained and communication (especially underwater) with other sensors/computers is near\hyp impossible. However, it is common practice to mount a GoPro (or other cameras) externally on different vehicles to provide a robot point view, with the most notable example being the Sharkcam from the Woods Hole Oceanographic Institute~\cite{sharkcam}. During our work on underwater cave mapping a GoPro 9 was attached to a custom made sensor suite~\cite{RahmanOceans2018} (termed stereo\hyp rig) containing a lower resolution stereo camera, IMU (including magnetometer), water depth sensor, and a mechanical scanning pencil\hyp beam sonar~\cite{sonar}. In this paper we will discuss two different post\hyp experiment synchronization methods between the GoPro and the stereo\hyp rig. In addition, the magnetometer data from the stereo\hyp rig are fused with the GoPro data for improving the accuracy of the state estimation process~\cite{RahmanIROS2019a}.}

Experiments conducted over an artificial reef (refuelling barge wreck) off the coast of South Carolina; inside the Devil System cave, FL; at the spring waters (open water) of Troy Springs State Park, FL; and in the cavern of Ginnie Springs, FL. In particular inside the Ginnie Springs cavern, five fiducial markers were placed in different locations in order to estimate the accuracy of the estimated trajectory. These datasets are made publicly available together with calibration data and basic scripts for evaluating ground truth\footnote{\url{https://afrl.cse.sc.edu/afrl/resources/datasets/}}. 

\invis{The next section discusses related work. An overview of the proposed setup is discussed in Section \ref{sec:setup}. The different datasets collected are outlined in Section \ref{sec:datasets}. Experimental results are presented in Section \ref{sec:results}. The paper concludes with a discussion on lessons learned and directions on current and future work.}

\section{Related Work}
A stereo GoPro setup was used to map underwater caves~\cite{WeidnerICRA2017}, however, this technology is no longer available, and it is only recently that the IMU of the GoPro 9 allows scale\hyp accurate results from a single camera. 
GoPro cameras have been studied in underwater settings~\cite{helmholz2019assessment,nocerino2019comparison}, and used, due to the high quality imagery in a variety of underwater tasks, for coral monitoring~\cite{nocerino2019comparison,neyer2019image,guo2016accuracy,gintert2012third}, underwater archaeology~\cite{van2015computer}, and seafloor reconstruction~\cite{raoult2016gopros,schmidt2012measurement}.  

Wreck mapping has been studied using a variety of techniques all around the world. Photogrammetry of manually obtained images resulted in mosaics in Demesticha \etal~\cite{demesticha20144th}, or from an ROV, see Nornes \etal~\cite{nornes2015underwater}. While the Arrows EU project provides an overview of robotic technology used~\cite{allotta2015arrows}. Menna \etal~\cite{menna2018state} provide a comprehensive review of techniques used. Mapping projects extend from Italy~\cite{balletti2015underwater}, Spain~\cite{palomeras2018autonomous}, Canada~\cite{coleman2002underwater}, Qatar~\cite{bain2014wreck}, up to the arctic~\cite{mogstad2020mapping}. With the most famous wreck explorations of the Titanic~\cite{eustice2006visually} and the Antikythera~\cite{williams2016return} shipwrecks.

Coral reef mapping also utilizes vision. By creating specialized sensors~\cite{hogue2006development,RahmanOceans2018} or utilizing UAVs~\cite{dunbabin2007large} there is a need for Underwater SLAM~\cite{williams2004simultaneous}. Due to the deteriorating health of the coral reefs, it is important to document the state of the different reefs and to measure the rate of deterioration. Of particular interest is to identify resilient species to assist re\hyp population efforts. 

There are few datasets from underwater experiments~\cite{QuattriniLiISERVO2016,JoshiIROS2019,ferrera2019aqualoc}, however, obtaining ground truth is extremely challenging. The Aqualoc dataset~\cite{ferrera2019aqualoc} used the trajectory estimated using global optimization package Colmap~\cite{colmap_sfm} as ground truth. This work will contribute a novel collection of datasets, and in select cases a set of permanent landmarks to act as ground truth. This dataset contains high definition/resolution images and inertial data at 200 Hz. The image quality is far superior than existing datasets. 

Mapping underwater caves is extremely challenging due to the total lack of ambient light. Wakulla Springs cave is one of the most well known and efforts to map it include the Wakulla 2 project~\cite{am20013d,stone2000automated} utilizing mainly acoustic sensors, as was mapping a cenote~\cite{gary20083d}. Nocerino \etal~\cite{nocerino2018multi} proposed the use of multiple cameras on a ROV for mapping caves, then use it to map caves in Sicily~\cite{nocerino20193d}. Malios \etal~\cite{mallios2016toward} proposed also a SLAM framework for confined spaces. The works of Rahman \etal~\cite{RahmanIROS2019a,RahmanICRA2018} has demonstrated accurate results over long trajectories in a variety of settings. On this work we utilize a subset of this work SVIn2 utilizing the initialization and loop closure extension over OKVIS~\cite{leutenegger2015keyframe}. Furthermore, a framework for denser reconstructions used the plethora of shadows in the cave environment~\cite{RahmanIROS2019b}. We have augmented this framework to produce a consistent map, by updating the triangulated features after every loop closure.

\section{Proposed Approach}
\label{sec:setup}
\subsection{Sensor Setup}
The GoPro 9  consists of a color camera, an IMU, and a GPS. GPS does not work underwater; thus it was not used in this work. However, GPS information can be fused with Visual Inertial Navigation Systems (VINS) during above\hyp water operations. For calibrating the camera intrinsic parameters and the extrinsic parameters of the sensor setup, we use a grid of AprilTags \cite{april_tags}. \tab{tab:gopro_teardown} presents the available sensors.

\begin{table}[h!]
\centering
{
\vspace{0.1in}
\resizebox{\columnwidth}{!}{
\begin{tabular}{@{}lccc}

\toprule
  Sensor & Type & Rate & Characteristics\\
 \midrule
 Camera & Sony IMX677  &  60 Hz & max. 5599$\times$4223, RGB color mosaic filters\\
 IMU & Bosch BMI260 & 200 Hz & 3D Accelerometer \& 3D Gyroscope \\
 GPS & UBlox UBX-M8030-CT & 18 Hz & 2 m CEP Accuracy\\  
\bottomrule
\end{tabular}}}
\caption{Overview of sensors in GoPro9 camera.}
\vspace{-0.1in}
\label{tab:gopro_teardown}
\end{table}

The GoPro 9 is equipped with Sony IMX677, a diagonal 7.85mm CMOS active pixel type image sensor with approximately 23.64M active pixels\invis{ and a maximum resolution of 5599$\times$4223}. GoPro 9 can run at 60 Hz at the maximum resolution of 4K. The Sony IMX677 sensor has an inbuilt 12-bit A/D converter to shoot high-speed and high-definition videos using horizontal and vertical binning and subsampling readout. The sensor has on-chip R, G, and B primary color mosaic filters for better color capture.  
GoPro 9 has multiple settings for recording the video, while many can be used there are certain modes which are prohibitive to VIO operations due to the non\hyp linear transformation of the image as detailed in \cite{goproModes}. We found that the SuperView mode generates non-linear distortions that thwart calibration of the camera intrinsic parameters during data collection. The videos were recorded at full High Definition (HD) resolution of 1960$\times$1080 with wide lens setting: horizontal field-of-view (FOV) 118\degree, vertical FOV 69\degree, and hypersmooth level set to off. Hypersmooth levels control the electronic image stabilization that predicts camera motion and compensates for it by cropping the view\hyp able image. Hypersmoothing can effectively crop up to 10\% of the image frame and the amount of cropping depends on the amount of motion, rendering this mode extremely challenging for VIO applications.

GoPro 9 also includes a Bosch BMI260 IMU equipped with 16-bit 3-axis MEMS accelerometer and gyroscope. GoPro 9 inherently records IMU data at 200Hz. The timestamps of IMU and camera are synchronized using the timing information from metadata encoded inside the MP4 video.

GoPro 9 also includes a UBlox UBX-M8030 GNSS chip capable of concurrent reception of up to 3 GNSS (GPS, Galileo, GLONASS, BeiDou) and accuracy of 2 m horizontal circular error probable, meaning 50\% of measurements fall inside circle of 2 m. \invis{Reception from more than one constellation simultaneously allows extraordinary positioning accuracy even with weak signals and high dynamics. GoPro 9 records GPS measurements at $\approx$18 Hz.}

\subsection{GoPro Telemetry Extraction}
The GoPro 9 MP4 video file is divided into multiple streams namely video encoded with H.265 encoder, audio encoded in advanced audio coding (AAC) format, timecode (audio-video synchronization information), GoPro fdsc data stream for file repair and GoPro telemetry stream in GoPro Metadata Format referred as GPMF\cite{gopro_parser}. GPMF -- is a modified Key, Length, Value solution, with a 32-bit aligned payload, that is both compact, fully extensible, and somewhat human readable in a hex editor. Please refer to~\cite{gopro_parser}  for more details on GPMF, here we focus on the camera-IMU synchronization.

\begin{figure}
\vspace{0.1in}
\centering
\includegraphics[width=0.55\columnwidth]{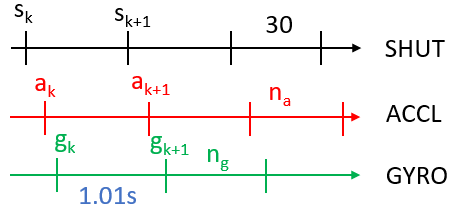}
\caption{Payload structure of GoPro metadata format with each payload containing 30 shutter exposure times, n$_a$ accelerometer measurements and n$_g$ gyroscope measurements in 1.01s at frame rate of 29.97Hz.}
\label{fig:gopro_payload}
\vspace{-0.1in}
\end{figure}

GPMF is divided into payloads, extracted using \textit{gpmf-parser}~\cite{gopro_parser}, with each payload containing sensor measurements for 1.01 seconds while recording at frame rate of 29.97 Hz as shown in \fig{fig:gopro_payload}. A particular sensor information is obtained from payload using FourCC--7-bit 4 character ASCII key, for instance 'ACCL' for accelerometer, 'GYRO' for gyroscope, and 'SHUT' for shutter exposure times. The payload also contains the starting time of each payload in microseconds relative to the start of the video capture. Since images are encoded in the video stream, we use the start of 'SHUT' payload to find relative timing with the accelerometer and gyroscope measurements. Using the start and end of payloads along with the number of measurements in that payload, we interpolate the timing of all measurements. We decode the video stream and extract images from MP4 file using the FFmpeg library and combine them with the IMU measurements using timing information from the GPMF payload.

\begin{figure}[ht]
\vspace{-0.1in}
\begin{center}
    \begin{subfigure}{0.49\columnwidth}
    \includegraphics[width=\textwidth, trim={0.17in 0in 0.3in 0.1in}, clip]{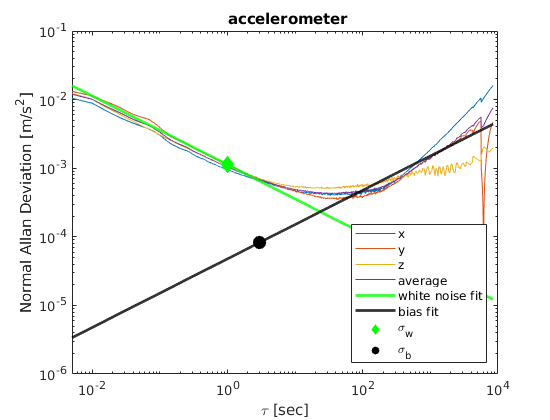}
    \end{subfigure}
    \begin{subfigure}{0.49\columnwidth}
    \includegraphics[width=\textwidth, trim={0.15in, 0in, 0.37in, 0.1in}, clip]{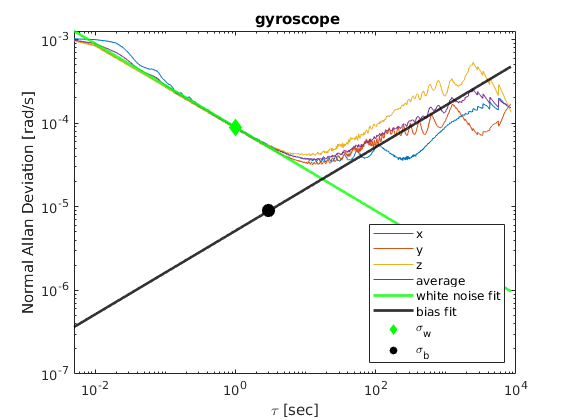}
    \end{subfigure}
    \caption{Allan deviation plot of accelerometer (left) and gyroscope (right) where we fit line with slope -1/2 and 1/2 averaged over all dimensions obtained using \cite{kalibr_allan}. The white noise $\sigma_w$ can be found at $\tau=1s$ on straight line with slope -1/2 and bias instability $\sigma_b$ at $\tau=3s$ on straight line with slope 1/2.}
    \label{fig:IMUplots}
\end{center}
\vspace{-0.1in}
\end{figure}

\subsection{Calibration}
Firstly, we calibrate the camera intrinsic parameters. We use the camera-calib sequences where we move the GoPro 9 camera in front of a calibration pattern placed in an indoor swimming pool from different viewing angles.  

The intrinsic noise parameters of the IMU are required for the probabilistic modeling of the IMU measurements used in state estimation algorithms and the camera-IMU extrinsic parameters. We assume that IMU measurements (both linear accelerations and angular velocities)  are perturbed by zero-mean uncorrelated white noise with standard deviation $\sigma_w$\ and random walk bias, which is the integration of white noise with standard deviation $\sigma_b$. To determine the characteristics of the IMU noise, the Allan deviation plot $\sigma_{Allan}(\tau)$ as a function of averaging time needs to be plotted. In a log-log plot white noise appears on Allan Deviation plot as a slope with gradient $-\frac{1}{2}$ and $\sigma_w$ can be found at $\tau = 1$. Similarly, the bias random walk can be found by fitting a straight line with slope $\frac{1}{2}$ at $\tau=3$; for details see~\cite{tumvio}. \fig{fig:IMUplots} shows the Allan deviation plot for GoPro 9 camera along with the noise parameters.

\subsection{Global Map}
Most VIO packages (\cite{RahmanIROS2019a,openvins,vinsmono}), when applying loop closure they update only the pose graph, leaving the triangulated features in their original estimate. In contrast COLMAP~\cite{colmap_sfm} is a global optimization package and the final result optimizes both poses and feature 3D locations. We run COLMAP by using 2 images per second resulting in 2000 to 3000 images in cavern and cave sequences, which takes on average 7 to 10 hours. ORB\hyp SLAM3~\cite{orbslam3} performs global bundle adjustment after loop closure, however often diverges over large dataset as the local mapping is stopped during global optimization. 

An enhancement is proposed for the SVIn2 framework to update the 3D pose of the tracked features every time loop closure occurs in order the 3D features to be consistent with the pose-graph optimization results. We maintain the pose graph with the keyframes as vertices for loop closure and the edges indicate the relative pose constraint between  keyframes. For each keyframe $f$, the VIO module passes the following information to the loop closure module:
\begin{itemize}
    \item $\boldsymbol{T}_{wf}$ pose of keyframe in world coordinate system
    \item 3D points visible in keyframe $l_i$ with each point $l = [ P_{w}, F, Q, I ]$ has the following attributes: position in world frame $P_{w} \in \mathbb{R}^3$, index of keyframe $F \in \mathbb{N}$, landmark quality $Q \in \mathbb{R}$, and position of keypoint in image $I_l \in \mathbb{N}^2$.
\end{itemize}

SVIn2, provides a quality measure of a 3D point as the ratio of the square root  of minimum and maximum eigenvalues of the Hessian block matrix associated with the 3D point. For each 3D point, we calculate its local position in keyframe as $P_{f} = \boldsymbol{T}^{-1}_{wf} P_{w}$, color $C$ as RGB value at pixel location $I$. We collect all the observations of 3D landmark $l$ from multiple keyframes as $O_f = [P_{f}, F, Q_f, C_f]$ hashed using keyframe index resulting in $O(1)$ lookup time for each observation. In the event of loop closure, we deform the global map so that the relative pose between each point and its attached keyframes remains unchanged. We fuse the multiple landmark observations from keyframe f = 1 to N to obtain global position $P_{w}$, color $C$ and quality $Q$ as shown in \eq{eq:weighted mean equation}.
\begin{equation}
\begin{aligned}
    P_{w} = &\frac{\sum_{f=1}^N T_{wf} * P_{f} * Q}{\sum_{f=1}^N Q}, C = \frac{\sum_{f=1}^N C_{f} * Q}{\sum_{f=1}^N Q}, \\
    & Q = \frac{\sum_{f=1}^N Q}{N}
\label{eq:weighted mean equation}
\end{aligned}
\end{equation}
\begin{figure}[h]
    \begin{center}
    \begin{tabular}{ll}        
    \begin{subfigure}{0.43\columnwidth}
        \fbox{\includegraphics[width=\textwidth,trim={0in 0in 0in 2.7in},clip]{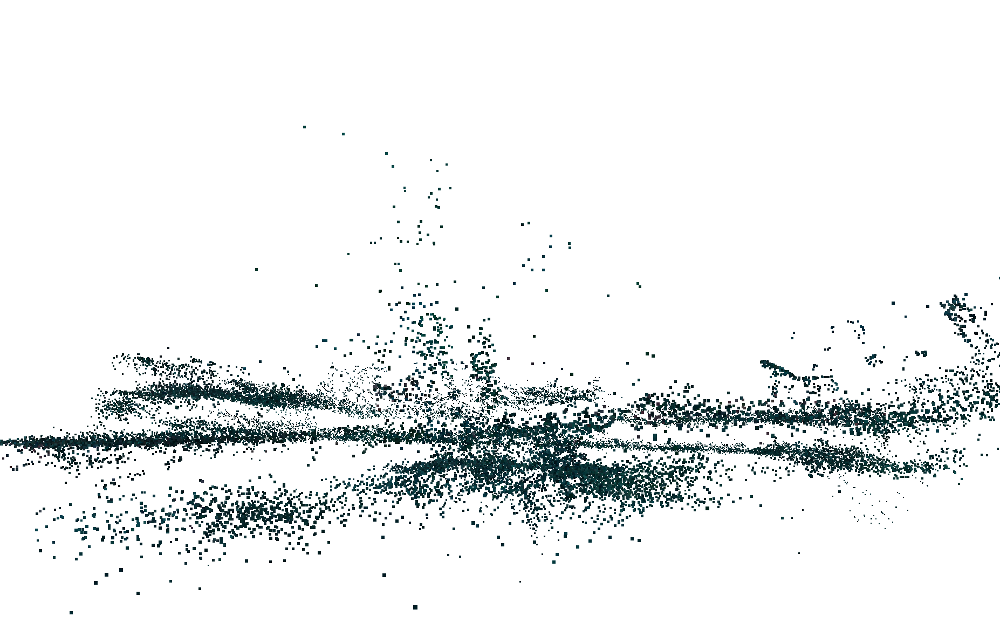}}
    \end{subfigure}
    \hspace{0.08in}
    \begin{subfigure}{0.43\columnwidth}
          \fbox{\includegraphics[width=\textwidth,trim={0in 0in 0in 2.7in},clip]{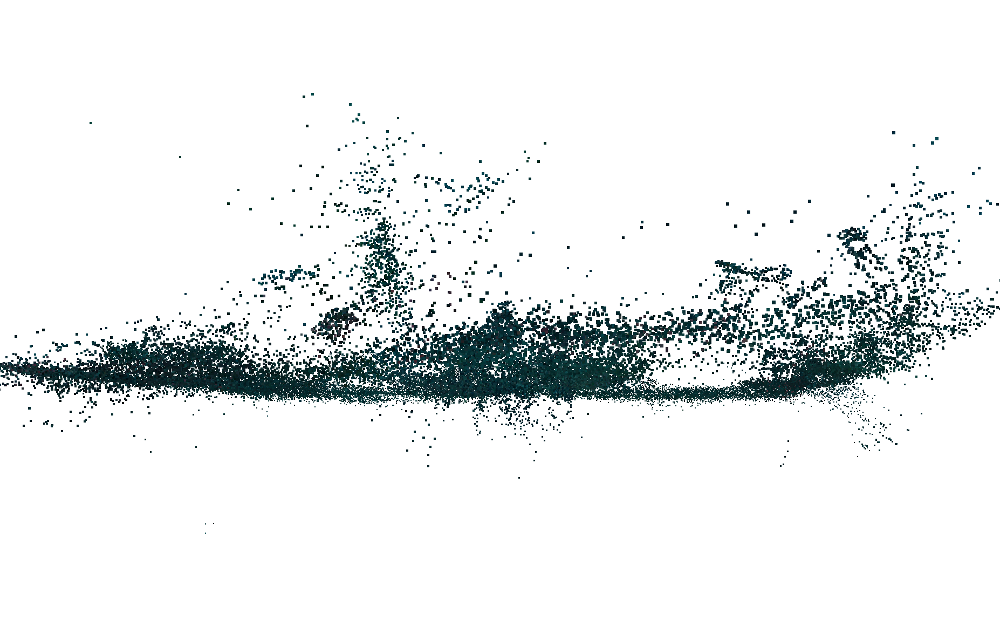}}  
    \end{subfigure}
    
    \end{tabular}
    \vspace{-0.05in}
    \caption{Feature points generated from SVIn2~\cite{RahmanIROS2019a} and with updated positions after loop closure. View of the X-Z plane, from the shipwreck\_1 sequence. Please note the three layers in the left image introduced by the gradual drift of the VIO process. Loop closure ensures that the z coordinates are consistent.}
    \label{fig:loop_pointclouds}
    \end{center}
    \vspace{-0.05in}
\end{figure}

\begin{figure*}[h]
    \begin{center}
    \leavevmode
    \captionsetup[subfigure]{aboveskip=-10pt, belowskip=-5pt}
    \begin{tabular}{lll}        
    \begin{subfigure}{0.24\textwidth}
        \includegraphics[width=\textwidth]{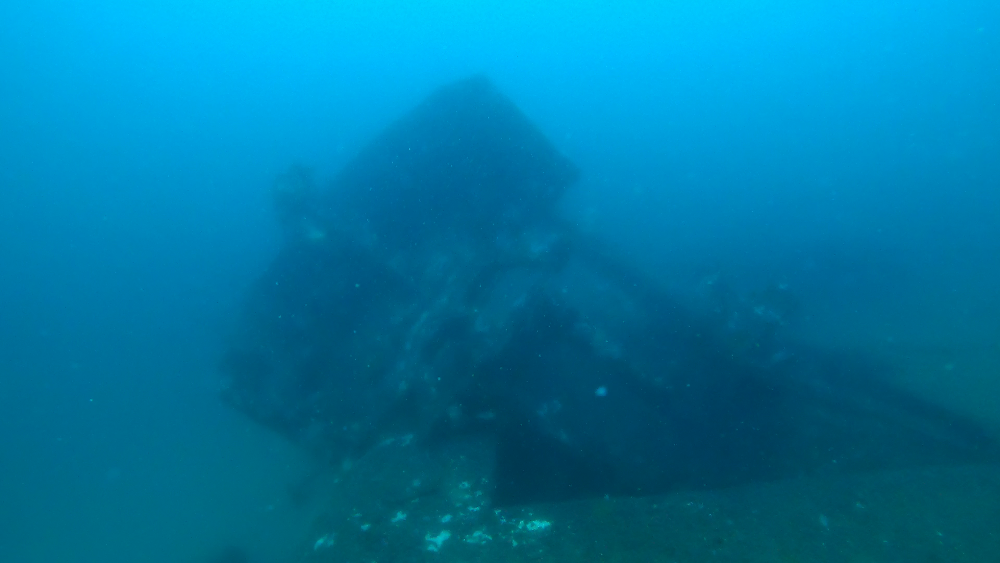}
        \label{l1}
        \caption{}
    \end{subfigure}

    \begin{subfigure}{0.24\textwidth}
        \includegraphics[width=\textwidth]{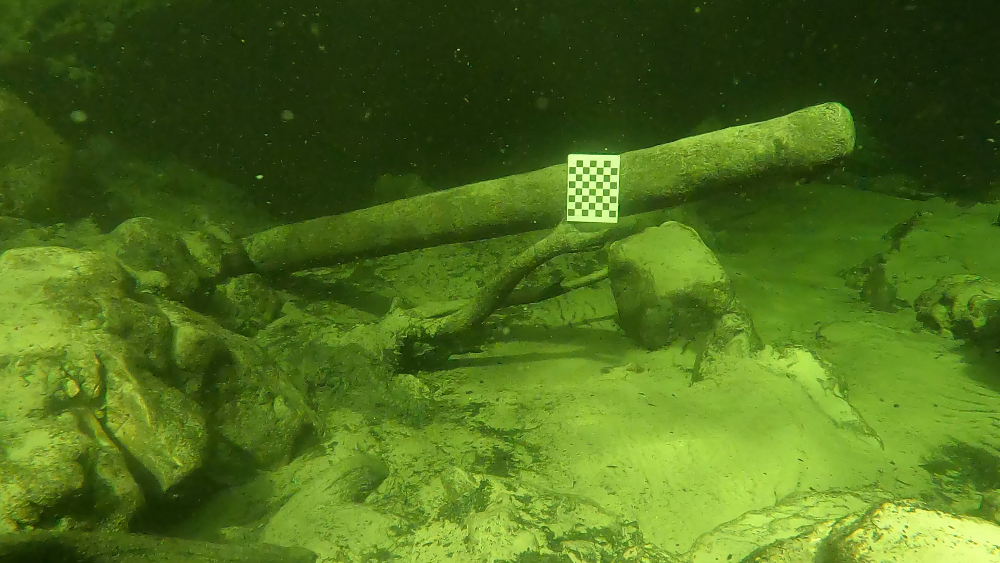}
        \label{l2}
        \caption{}
    \end{subfigure}
    
    \begin{subfigure}{0.24\textwidth}
        \includegraphics[width=\textwidth]{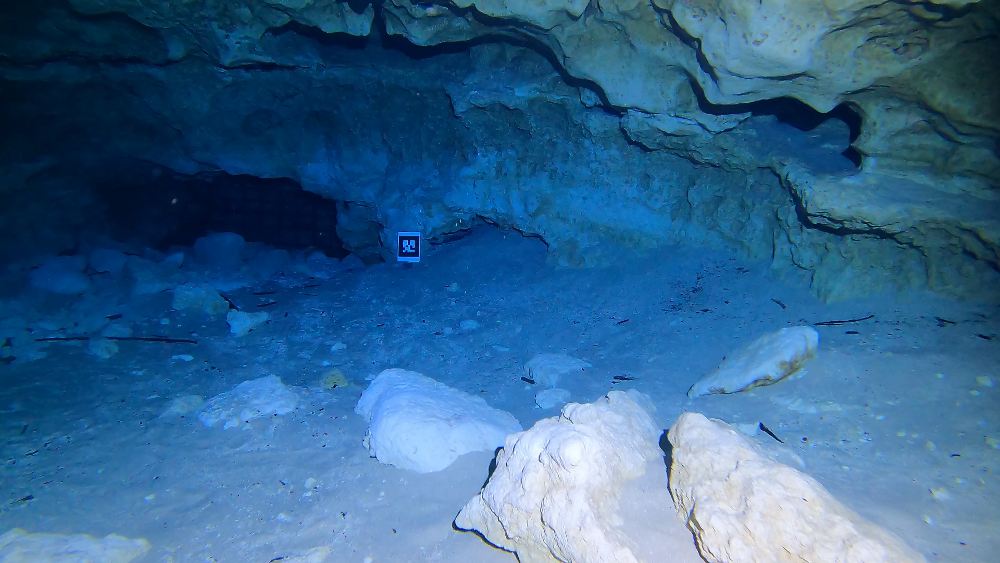}
        \label{l3}
        \caption{}
    \end{subfigure}

    \begin{subfigure}{0.24\textwidth}
        \includegraphics[width=\textwidth]{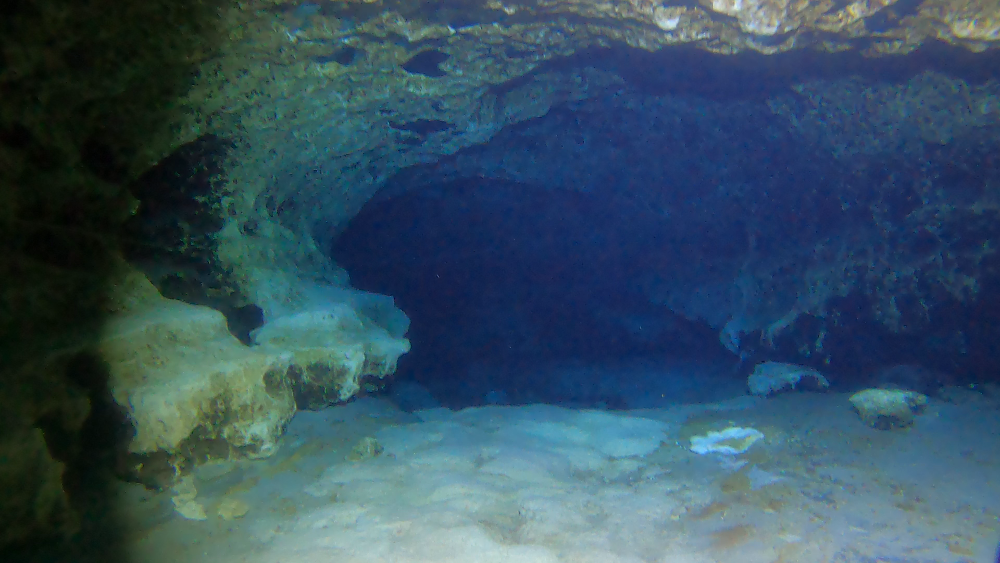}
        \label{l4}
        \caption{}
    \end{subfigure}

    \end{tabular}
    \caption{Images from the different environments: (a) Shipwreck; (b) Troy Springs; (c) Ginnie Springs cavern; (d) Devil's system cave.}
    \vspace{-0.1in}
    \label{fig:snapshots}
    \end{center}
\end{figure*}

It should be noted that $P_f$ is expressed as the relative position with respect to the keyframe and whenever the pose of keyframe $T_{wf}$ changes due to loop closure updates, the location of the 3D landmarks in the global frame also changes accordingly; producing a globally consistent map as shown in \fig{fig:loop_pointclouds}. Realistically, the number of observations of a 3D point is quite small compared to the total 3D points in the scene; thus global map building scales linearly $O(n)$ with the number of points in the scene.
\invis{
\subsection{External Synchronization}
When the GoPro is rigidly attached to another device (AUV, ROV, sensor suite) there are two different methods to synchronize the two data streams: a) based on global lighting variations; b) based on drastic orientation changes with respect to the gravity vector. 
\begin{enumerate}[label=\alph*]
    \item When illumination is provided by an artificial source under the control of the operator, then by turning off and on the lights a couple of times, the average intensities of the two vision streams can be calculated and the offset of the two signals identified. However, if the only source of illumination is ambient light, then the only option is to try to cover the two cameras which makes the synchronization more difficult.
    \item The IMU data-streams underwater are dominated by the gravity acceleration. When the sensors are left at rest, and when the orientation has large changes in roll and pitch, the roll and pitch signals can be matched between the two IMUs in order to synchronize the clocks of the two devices.
\end{enumerate}
In this work we have used the IMU synchronization between the GoPro 9 and the stereo\hyp rig. Consequently, the magnetometer data from the stereo\hyp rig are integrated with the visual/inertial data of the GoPro.

\subsection{Magnetometer based Corrections}
}

\section{Datasets}
\label{sec:datasets}
The GoPro9 Underwater VIO dataset consists of calibration sequences in addition to odometry evaluation sequences recorded using 2 GoPro 9 cameras. The two cameras are referred as g$_i, ~i \in [1,2]$. Although we provide calibration results for all the datasets, the calibration sequences are made available to facilitate users wanting to preform their own calibration. Datasets can be categorized as:

\begin{itemize}
    \item \textbf{camera-calib:} for calibrating the camera intrinsic parameters underwater. We provide calibration sequences using two calibration patterns:  a grid of AprilTags and a checkerboard for each camera. The calibration patterns are recorded with slow camera motions  at a swimming pool; making sure the calibration patterns are viewed from varying distances and orientations.
    
    \item \textbf{camera-imu-calib:} for calibrating the camera-IMU extrinsic parameters in order to determine the relative pose between the IMU and the camera. The camera is moved in front of the Apriltag grid exciting all 6 degrees of freedom. This sequence is recorded indoor just to expedite the calibration process as the extrinsic parameters are the same above and below water. Moreover, we also provide a \textbf{camera-calib} indoor sequence to assist with the camera-IMU calibration.
    
    \item \textbf{imu-static:} contains IMU data to estimate white noise and random walk bias parameters. These sequences are recorded with the camera stationary for at least 4 hours for each cameras. 
    
    \item \textbf{shipwreck:} two sequences collected by handheld GoPro 9 on an artificial reef (refueling barge wreck) ~55 Km outside of Charleston, SC, USA; see \fig{fig:snapshots}(a).
    
    \item \textbf{spring open water} one sequence collected by handheld GoPro 9 at the basin of Troy Springs State Park FL, USA; see \fig{fig:snapshots}(b). The settings had image stabilization on (hypersmooth) resulting into arbitrary cropping of the field of view. 
    
    \item \textbf{cavern:} three trajectories traversed inside the ballroom cavern at Ginnie Springs, FL; see \fig{fig:snapshots}(c). Inside the cavern five markers (AR single tags~\cite{ar_tags}) were placed to establish ground truth measurements. Each trajectory consisted of several loops each observing slightly different parts of the cavern but all ensuring the tag of that part of the cavern was visible. 
    
    \item \textbf{cave:} two sequences were collected at the Devil's system, FL; see \fig{fig:snapshots}(d). For the first sequence the GoPro was mounted on the stereo rig and second sequence was using only the GoPro.    
\end{itemize}

\section{Experimental Results}
\label{sec:results}
Due to absence of GPS in underwater environments or motion capture systems, we use COLMAP \cite{colmap_sfm} to generate baseline trajectories. COLMAP is a structure-from-motion (SfM) pipeline equipped with global bundle adjustment and loop closure capabilities; thus producing consistent camera trajectory and 3D reconstruction. Even though COLMAP provides good estimation of shape of trajectories, they can not be considered as ground truth. As monocular SfM inherently suffers from scale observability constraints and global optimization does not converge over large trajectories, we consider the estimated trajectories as accurate up to scale. We found that relative scale between COLMAP and all VIO trajectories was almost equal. Hence, COLMAP trajectories are scaled by scaling factor calculated from the average of all VIO trajectories in subsequent sections unless otherwise specified.

\subsection{Tracking Evaluation Metrics}
As COLMAP does not provide accurate scale, we evaluate the accuracy of the various tracking algorithms using the \textit{absolute trajectory error (ATE)} metric after \textbf{Sim(3)} alignment \cite{Umeyama:1991:LET:105514.105525}. ATE is calculated as the root mean squared difference between ground truth 3D positions obtained from COLMAP $\boldsymbol{p}_i$ and corresponding estimated 3D positions $\boldsymbol{\hat{p}}_i$ aligned using optimal \textbf{Sim(3)}  rotation matrix $\boldsymbol{R}$, translation $\boldsymbol{t}$, and scaling factor $\boldsymbol{s}$:

\begin{equation}
    e_{ATE} =\min_{\boldsymbol{(R,t,s)}\in \textbf{Sim(3)}} \sqrt{\frac{1}{n} \sum_{i=1}^{n} \parallel \boldsymbol{p}_i - (\boldsymbol{s R \hat{p}_i + t}) \parallel^2}
\end{equation}

\subsection{Tracking Results}
We compare the performance of various open source visual-inertial odometry (VIO) methods on the above described datasets. Since, most of the VIO algorithms have parameters that are tuned at VGA resolution; their performance was found to be better at quarter resolution (960$\times$540 pixels). However, we provide datasets at full high definition resolution (1920$\times$1080 pixels) to further the research in underwater VIO and SfM. Unless specified otherwise, all the VIO algorithms use quarter resolution dataset.  

We evaluate the performance of VINS-Mono \cite{vinsmono}, ORB\hyp SLAM3\cite{orbslam3}, SVIn2\cite{RahmanIROS2019a} and OpenVINS\cite{openvins} based on the RMSE of the ATE as shown in Table \ref{tab:rmse_error}. All the compared algorithms are equipped with loop closure resulting in low overall RMSE error.  It should be noted that OpenVINS is based on the Mutiple-State Constraint Kalman Filter, where as all other methods are based on non-linear least squares optimization.


\begin{figure*}
\vspace{-0.05in}
\centering
    \begin{subfigure}{0.25\textwidth}
        \includegraphics[width=\textwidth, trim={0.35in, 0.40in, 0.7in, 0.60in}, clip]{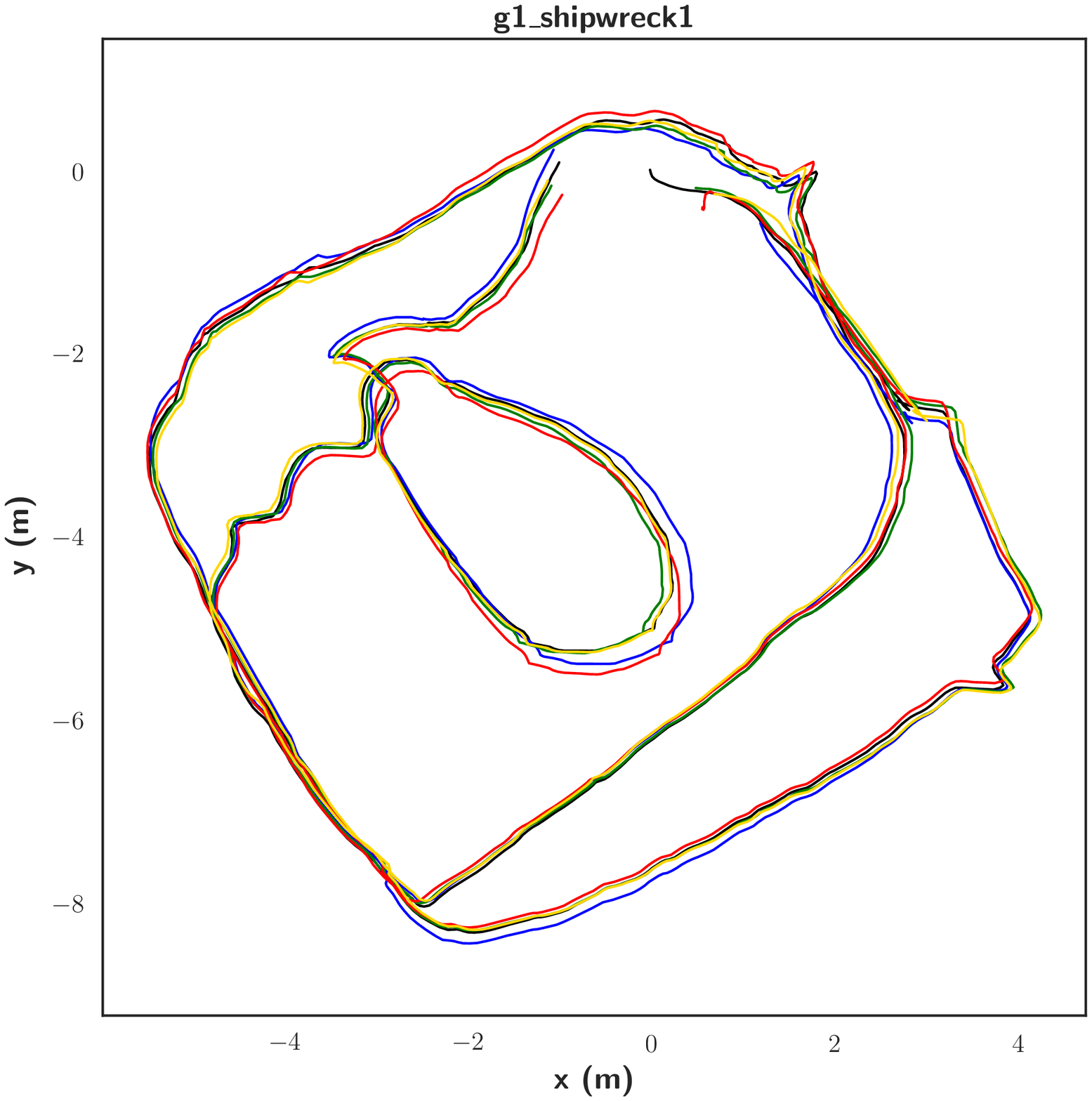}
    \end{subfigure}%
    \begin{subfigure}{0.25\textwidth}
        \includegraphics[width=\textwidth, trim={0.35in, 0.40in, 0.7in, 0.60in}, clip] {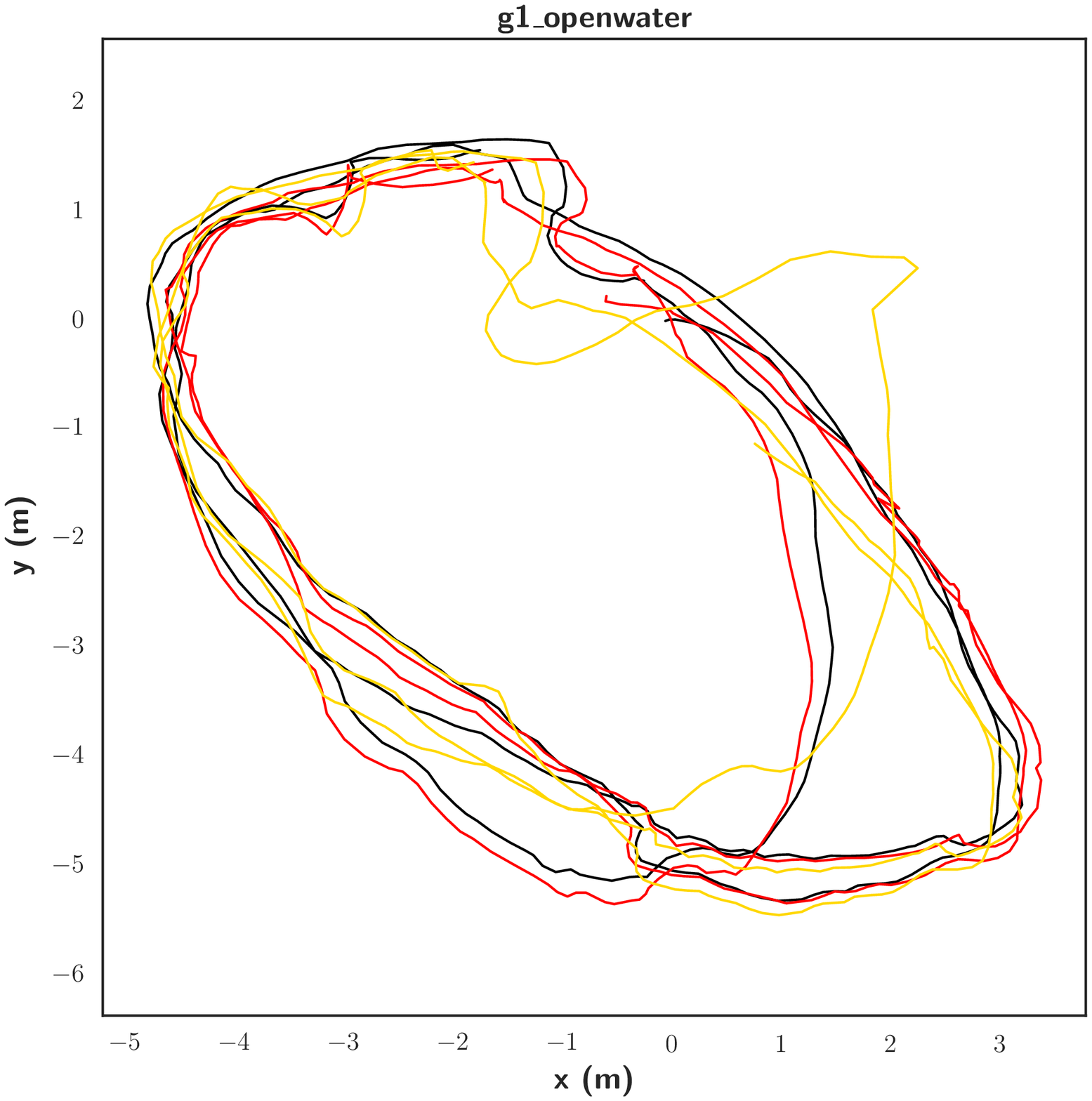}
    \end{subfigure}%
    \begin{subfigure}{0.25\textwidth}
        \includegraphics[width=\textwidth, trim={0.35in, 0.40in, 0.7in, 0.60in}, clip]{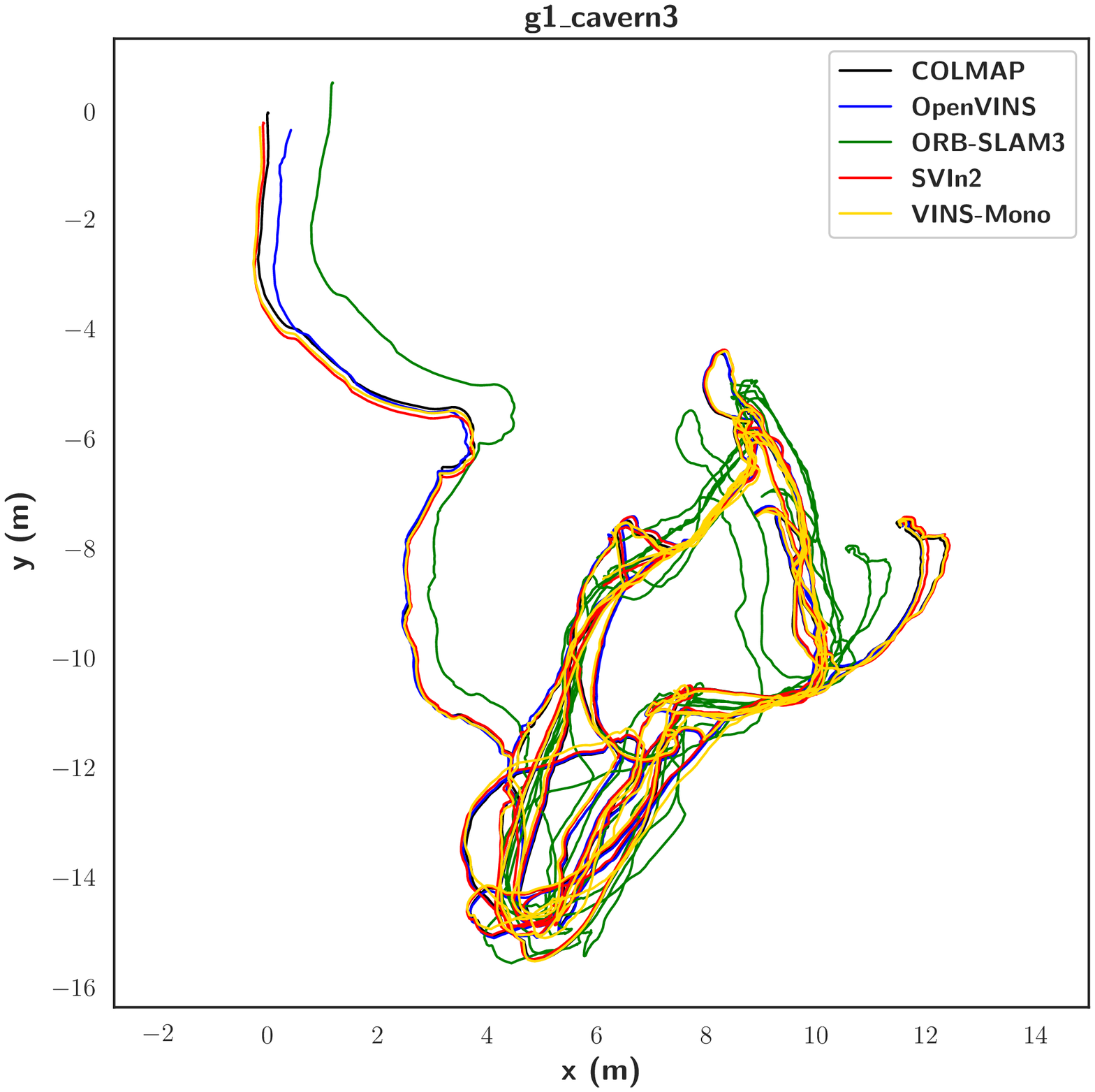}
    \end{subfigure}%
    \begin{subfigure}{0.25\textwidth}
        \includegraphics[width=\textwidth, trim={0.35in, 0.40in, 0.7in, 0.60in}, clip]{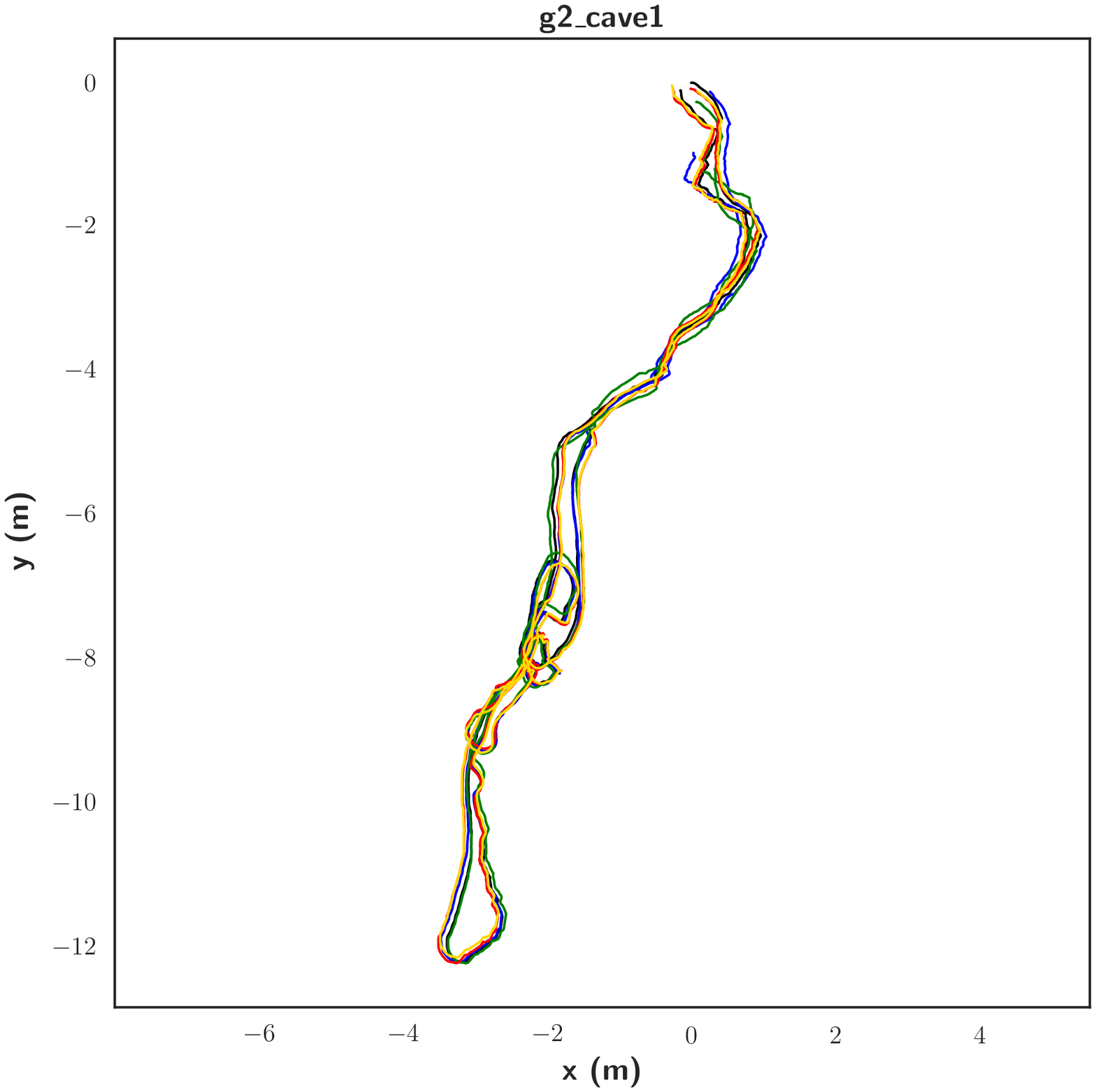}
    \end{subfigure}

\caption{Trajectories of the evaluated algorithms on g1\_shipwreck1, g1\_openwater, g1\_cavern3, and g2\_cave1 sequences from our GoPro Underwater dataset. COLMAP (black) obtained using global bundle adjustment is used as ground truth up to scale. All other trajectories are obtained by enforcing real time constraints; even if serial processing modes are available for ORB\hyp SLAM3 and OpenVINS.}
\label{fig:three graphs}
\end{figure*}

\begin{table}[h!]
\centering {
\begin{tabular}{@{}lccccc}
\toprule
 \rotatebox{90}{Sequence} &  \rotatebox{90}{length[m]} & \rotatebox{90}{SVIn2} & \rotatebox{90}{ORB\hyp SLAM3} & \rotatebox{90}{VINS-Mono} & \rotatebox{90}{OpenVINS} \\
 \midrule
 {g1\_shipwreck1} & {71.60} & {0.148} & \textbf{0.098}  &  0.143 & 0.167\\
{g1\_shipwreck2} & {79.07} & {0.238} & 0.207  & \textbf{0.202} & 0.337\\
 {g1\_openwater} & {53.26} & \textbf{0.263} & $\times$ & 0.486 & $\times$ \\
  {g1\_cavern1} & {243.10} & \textbf{0.090} & 0.231  & \textbf{0.090} & 0.288\\
  {g1\_cavern2} & {240.60} & \textbf{0.074} & 0.363  & 0.097 & 0.084\\
   {g2\_cavern3} & {341.29} & {0.089} & {0.687}  & {0.131} & \textbf{0.081}\\
  {g2\_cave1} & {219.58} & \textbf{0.072} & 0.150  & 0.105 & 0.081\\
   {g2\_cave2} & {222.64} & 0.093 & \textbf{0.089}  & 0.097 & 0.229\\
\bottomrule
\end{tabular}}
\caption{Performance of evaluated open source algorithms on various datasets based on root mean squared ATE in meters. }
\label{tab:rmse_error}
\end{table}

All system are able to track most of the sequences until the end. VINS-Mono and SVIn2 were able to track the complete trajectory consistently with good accuracy. In the cavern sequences, ORB\hyp SLAM3 took too long during global bundle adjustment after loop closure and lost track as it disabled local mapping during global optimization. However, ORB\hyp SLAM3 is equipped with map merging and was able to relocalize and merge the disjoint maps. This produced slightly inferior performance in the cavern sequences. OpenVINS required smooth motion for initialization as it only relies on IMU measurement for gravity alignment and orientation initialization. Hence, OpenVINS might diverge unless data collection is started from a static position or good initialization is found. Finally, it is worth noting that the open water sequence was recorded with the hypersmooth option activated resulting in failures in ORB\hyp SLAM3 and OpenVINS.

\subsection{AR-tag based Validation}
As there is no continuous tracking for absolute ground truth, 3D landmark-based validation with AR tags is used to quantify the accuracy of the evaluated methods. As a part of the experimental setup in the cavern sequences with multiple loops, we placed 5 different AR-tags printed on waterproof paper at different locations inside the cavern. We observe the variance of the position of the AR-tags from their mean position over the whole length of trajectory. If the trajectories do not drift over time, the markers must be observed at the same location during multiple visits. Among the cavern sequences, we detected most AR-tags in the g1\_cavern2 sequence; therefore, this sequence is used as reference for further analysis.

We determined the relative position between the camera and the tags in g1\_cavern2 sequence using \textit{ar\_track\_alvar}\footnote{\url{http://wiki.ros.org/ar\_track\_alvar}}. By projecting a 3D cube over the tags using the pose estimate, we observed higher noise in the orientation estimate; hence, only the position of the AR-tags is used for the error analysis. Once the relative position from \textit{ar\_track\_alvar} is found,  the global position can be found as $\boldsymbol{T}_{WM}^k = \boldsymbol{T}_{WC}^k * \boldsymbol{P}_{CM}$ where  $\boldsymbol{T}_{WM}^k$ is the marker position in world coordinate frame W at time k, $\boldsymbol{T}_{WC}^k$ is the pose of the camera C in W at time k (produced by SLAM/odometry system), and $\boldsymbol{P}_{CM}$ is the relative position of marker M from camera. 

\begin{figure}[ht]
\centering
\includegraphics[width=0.30\textwidth]{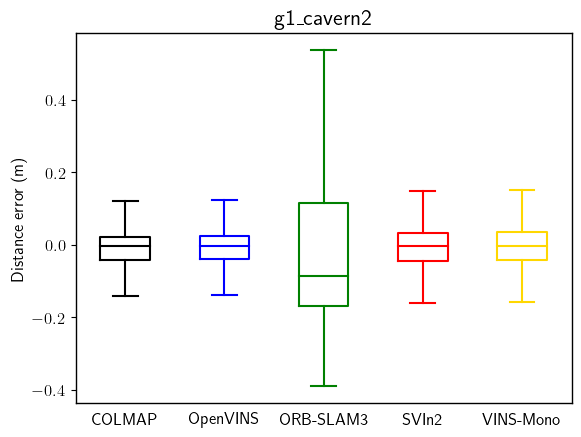}
\caption{Boxplot of distance error for all tags visible in g1\_cavern2.}
\label{fig:tag_boxplot}
\end{figure}

\begin{figure}[ht]
\centering
\includegraphics[width=0.35\textwidth]{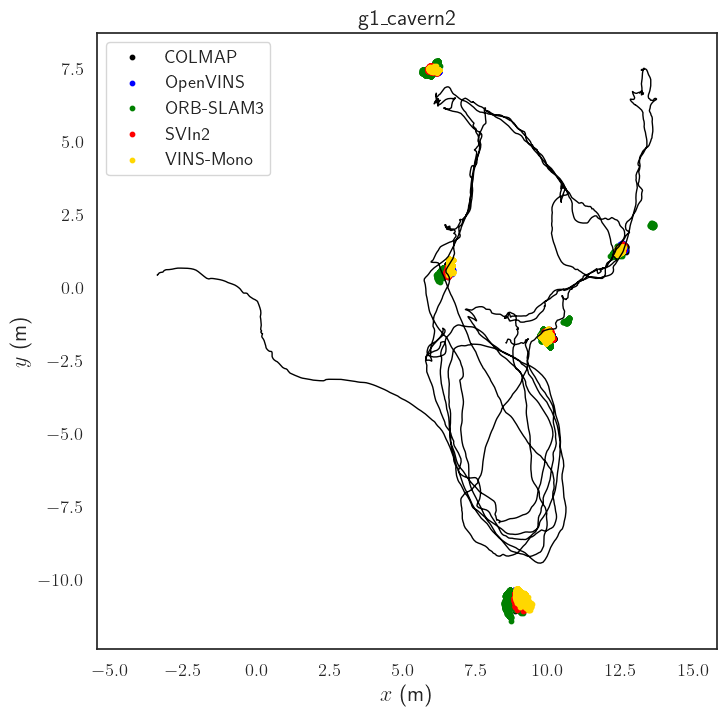}
\caption{The position of observed tags (in total 5) placed at different places inside the cavern as calculated using evaluated algorithms. The COLMAP trajectory also shown as reference.\\}
\label{fig:tag_plot}
\end{figure}

\fig{fig:tag_boxplot} shows boxplots of the displacement from the mean position of the markers over the whole length of the trajectory  for all the different methods, including COLMAP. \tab{tab:standard_deviation_translation} shows the summary of the  standard deviation in translation and the average distance error. All the algorithms performed well with slightly inferior performance of ORB\hyp SLAM3 due to tracking issues. \fig{fig:tag_plot} shows the position of the tags in g1\_cavern2 sequence observed by different packages along with the COLMAP trajectory as reference.

\begin{table}[h]
\centering
{
\begin{tabular}{@{}lcccc}

  &  $t_x $(m) & $t_y $(m) & $t_z $(m) & Avg dist. error (m)\\
 
 \toprule
 COLMAP & \textbf{0.057} & 0.072 & \textbf{0.034} & \textbf{0.099} \\
 OpenVINS & 0.063 & \textbf{0.071} & \textbf{0.034} & 0.104 \\
 ORB\hyp SLAM3 & 0.246 & 0.222 & 0.068 &  0.268 \\
 SVIn2 & 0.058 & 0.075 &  0.049 & 0.106 \\
 VINS-Mono & 0.073 & 0.086 & 0.045 & 0.122\\  
\bottomrule
\end{tabular}}
\caption{Standard deviation in translation of detected tags and average distance error.}
\label{tab:standard_deviation_translation}
\end{table}


\vspace{-0.03in}
\subsection{Global Mapping}
We evaluate the global map produced by enhancing SVIn2 to update the triangulated feature positions after loop closures by comparing with COLMAP's sparse pointcloud. The pointclouds come from different sources and differ in size, so we align the pointclouds from COLMAP and SVIn2 as follows:

\begin{enumerate}
    \item Perform voxel downsampling with voxel size of 10cm.
    \item Compute FPFH \cite{fpfh} feature descriptor describing local geometric signature for each point.
    \item Find correspondence between pointclouds by computing similarity score between FPFH descriptors.
    \item Feed all putative correspondences to TEASER++ \cite{teaser} to perform global registration finding transformation to align corresponding points.
    \item Fine tune registration by running ICP over original  point cloud with TEASER++ solution as initial guess.
\end{enumerate}

\begin{figure}
\centering
    \vspace{-0.2in}
    \begin{subfigure}{0.40\columnwidth}
        \includegraphics[width=\textwidth, trim={4in, 0in, 2in, 1.1in},clip]{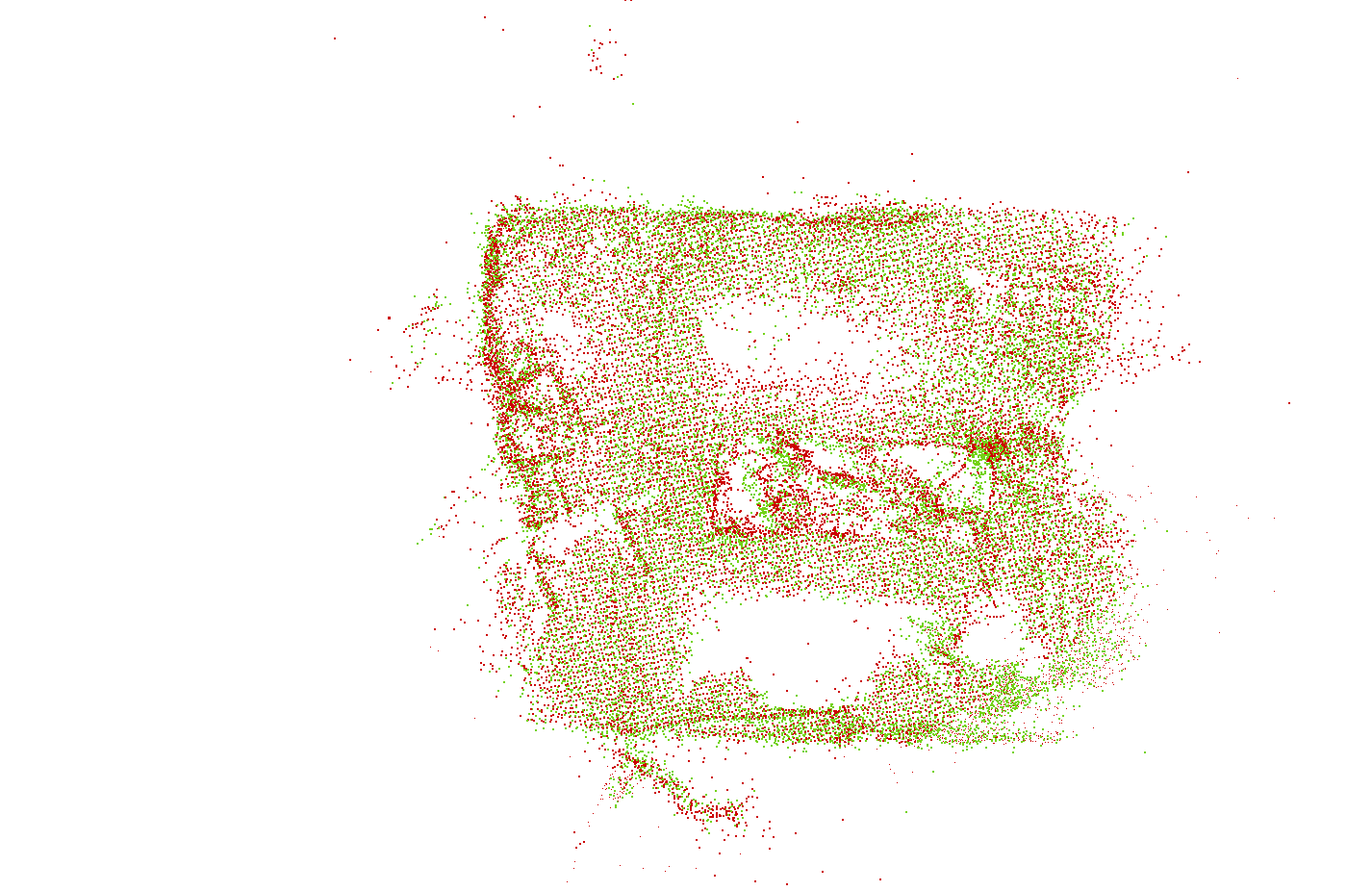}
    \end{subfigure}
    \begin{subfigure}{0.50\columnwidth}
        \includegraphics[width=\textwidth, trim={2in, 0in, 2in, 1.1in},clip]{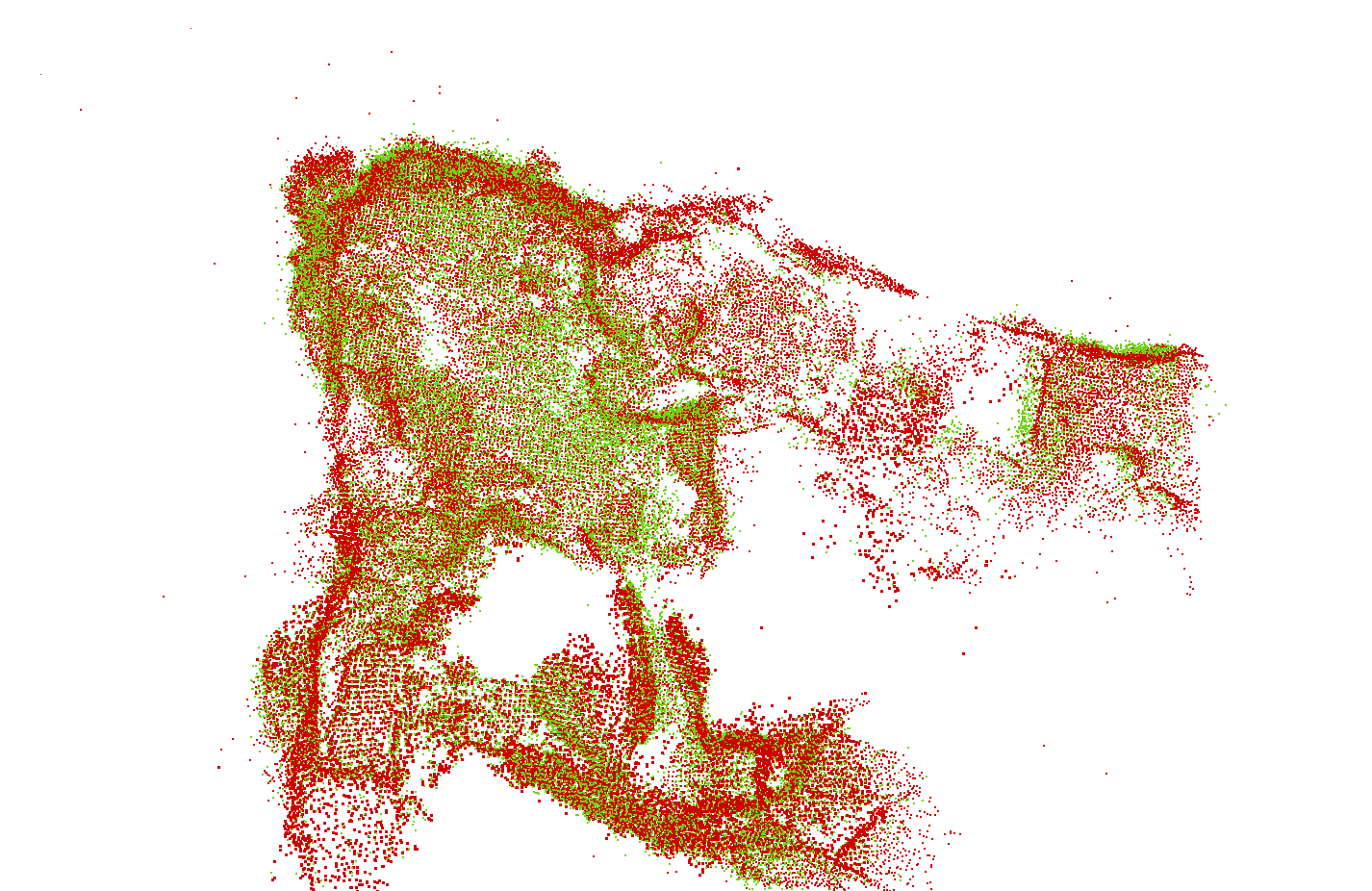}
    \end{subfigure}

    \caption{Sparse reconstruction obtained using COLMAP (red) and SVIn2 (green) from the g1\_shipwreck1 and g1\_cavern2 sequences. }
\label{fig:global_reconstruction}
\vspace{-0.2in}
\end{figure}

The reconstruction results are compared based on registration accuracy using fitness and inlier\_rmse metrics. More specifically, fitness is the ratio of number of inlier correspondences (distance less than voxel size) and number of points in SVIn2 pointcloud. Whereas, inlier\_rmse is the root mean squared error of all inlier correspondences. \tab{tab:reconstruction_result} shows the similarity between SVIn2 and COLMAP reconstruction based on fitness and inlier\_rmse metrics. \fig{fig:global_reconstruction} shows aligned sparse reconstruction obtained from COLMAP and SVIn2 in g1\_shipwreck1 and g1\_cavern2 sequence.

\begin{table}[h!]
\centering
{
\begin{tabular}{@{}lccr}

  &  fitness & inlier\_rmse & \# of correspondences\\
 
 \toprule
 g1\_shipwreck1 &  0.773 & 0.058 & \num{12391}\\
 g1\_shipwreck2 & 0.500 & 0.063 & \num{9959} \\
 g1\_cavern1 & 0.738 & 0.058 & \num{34481} \\
 g1\_cavern2 & 0.727 & 0.060 &  \num{44797} \\
 g2\_cavern3 & 0.698 & 0.059 & \num{46875} \\
 g2\_cave1 & 0.107 & 0.067 & \num{6302}\\
 g2\_cave2 & 0.326 & 0.064 & \num{19750} \\
\bottomrule
\end{tabular}}
\caption{Similarity between SVIn2 and COLMAP sparse reconstructions based on fitness and inlier\_rmse metrics along with no. of correspondences.}
\label{tab:reconstruction_result}
\end{table}

\begin{figure}[ht]
\vspace{-0.1in}
    \begin{center}
    \begin{subfigure}{0.60\columnwidth}
        \includegraphics[width=\textwidth]{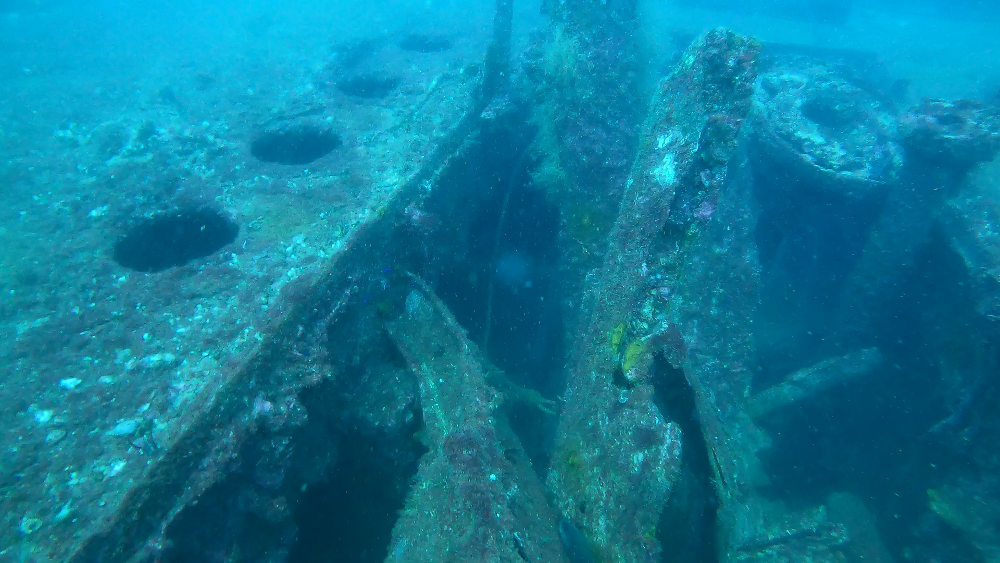}
    \end{subfigure}
    \begin{subfigure}{0.38\columnwidth}
        \includegraphics[width=\textwidth]{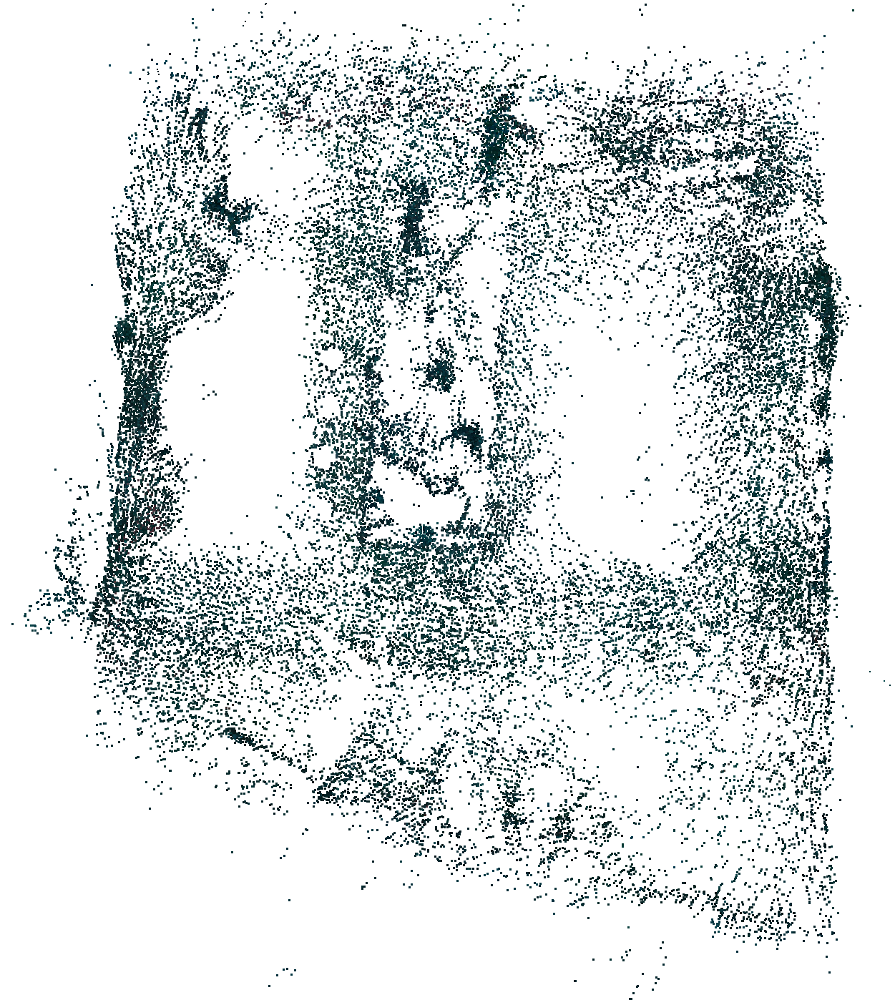}
    \end{subfigure}
    
    \vspace{-0.1in}
    \caption{A view from the GoPro showing part of the wreck's deck and the global map showing the 3D reconstruction of the mapped area. Please note the round in the reconstruction next to the middle opening with the broken beams.}
    \label{fig:GoProPTS}
    \end{center}
\end{figure}

\section{Conclusion}
In this work we presented a complete pipeline for underwater SLAM utilizing a commonly available, inexpensive, action camera with superior performance. The proposed approach was tested in open and confined waters, with natural and artificial illumination, under challenging conditions. The SVIn2 framework was augmented to correct the 3D features according to the updated pose graph after successful loop closure calculations. The resulting map demonstrated accuracy similar to the much slower global optimization COLMAP package. The experimental results verify that the specific camera is capable of producing accurate estimates of the trajectory together with consistent sparse representations of the environment. Future uses of the proposed framework would be in recording and documenting the surroundings and the trajectory of AUVs operating autonomously in challenging underwater environments such as caves and shipwrecks~\cite{XanthidisICRA2020, XanthidisIROS2021}.

Currently we are investigating synchronization methods between the GoPro camera and other devices such as Autonomous Underwater Vehicle and sensor suites. By introducing additional data streams such as water depth and magnetometer data, both providing absolute values, we expect to reduce the drift accumulating over long trajectories without loops and increase the overall accuracy.

\invis{
\section*{Acknowledgements}
The authors would also like to acknowledge the help of the Woodville Karst Plain Project (WKPP) and El Centro Investigador del Sistema Acuífero de Quintana Roo A.C. (CINDAQ) in collecting data, providing access to challenging underwater caves, and mentoring us in underwater cave exploration. Last but not least, we would like to thank Halcyon Dive Systems for their support with equipment.
}
\pagebreak
\newpage
\printbibliography

\end{document}